\documentclass{article}

\PassOptionsToPackage{numbers}{natbib}

\usepackage[final]{neurips_2024}

\usepackage[utf8]{inputenc} 
\usepackage[T1]{fontenc}    
\usepackage{hyperref}       
\usepackage{url}            
\usepackage{booktabs}       
\usepackage{amsfonts}       
\usepackage{nicefrac}       
\usepackage{microtype}      
\usepackage{xcolor}         
\usepackage{multirow}
\usepackage{xcolor,colortbl}
\usepackage{adjustbox}
\usepackage{siunitx}
\usepackage[font=small]{caption}
\usepackage{placeins}

\title{Atlas: A Novel Pathology Foundation Model by\\ Mayo Clinic, Charité, and Aignostics}

\author{
Maximilian Alber$^{\;*\;1\;12}$,
Stephan Tietz$^{\;*\;1}$, 
Jonas Dippel$^{\;*\;1\;6\;7}$,
Timo Milbich$^{\;*\;1}$,\AND
Timothée Lesort$^{\;*\;1}$,
Panos Korfiatis$^{\;\#\;3}$,
Moritz Krügener$^{\;\#\;1}$,
Beatriz Perez Cancer$^{\;\#\;1}$,\AND
Neelay Shah$^{\;\#\;1}$,
Alexander Möllers$^{\;1\;6\;7}$,
Philipp Seegerer$^{\;1}$,
Alexandra Carpen-Amarie$^{\;1}$,\AND
Kai Standvoss$^{\;1}$,
Gabriel Dernbach$^{\;1\;7\;12}$,
Edwin de Jong$^{\;1}$,
Simon Schallenberg$^{\;12}$,\AND
Andreas Kunft$^{\;1}$,
Helmut Hoffer von Ankershoffen$^{\;1}$,
Gavin Schaeferle$^{\;5}$,
Patrick Duffy$^{\;4}$,\AND
Matt Redlon$^{\;4}$,
Philipp Jurmeister$^{\;10\;11}$,
David Horst$^{\;10\;12}$,
Lukas Ruff$^{\;1}$,\AND
Klaus-Robert Müller$^{\dag\;6\;7\;8\;9}$,
Frederick Klauschen$^{\dag\;7\;10\;11\;12\;13}$,
Andrew Norgan$^{\dag\;2}$\\ \\
$^1$ Aignostics, Germany \AND 
$^2$ Department of Laboratory Medicine and Pathology, Mayo Clinic, Rochester, MN, US \AND
    $^3$ Department of Radiology,  Mayo Clinic, Rochester MN, US \AND 
    $^4$ Department of Information Technology, Mayo Clinic, Rochester MN, US \AND 
    $^5$ Systems Quality Office, Mayo Clinic, Rochester MN, US \AND 
    $^6$ Machine Learning Group, Technische Universität Berlin, Germany \AND 
    $^7$ BIFOLD – Berlin Institute for the Foundations of Learning and Data, Germany \AND 
    $^8$ Department of Artificial Intelligence, Korea University, Republic of Korea \AND 
    $^9$ Max-Planck Institute for Informatics, Germany \AND 
    $^{10}$ German Cancer Research Center (DKFZ) \& German Cancer Consortium (DKTK), \\ Berlin \& Munich Partner Sites, Germany \AND 
    $^{11}$ Institute of Pathology, Ludwig-Maximilians-Universität München, Germany \AND 
    $^{12}$ Institute of Pathology, Charité – Universitätsmedizin Berlin, Germany \AND 
    $^{13}$ Bavarian Cancer Research Center (BZKF), Germany \AND 
    $^{*, \#, \dag}$ Equal contribution respectively}

\begin{document}

\maketitle

\begin{abstract}
Recent advances in digital pathology have demonstrated the effectiveness of foundation models across diverse applications.
In this report, we present Atlas, a novel vision foundation model based on the RudolfV approach.
Our model was trained on a dataset comprising $1.2$ million histopathology whole slide images, collected from two medical institutions: Mayo Clinic and Charité - Universtätsmedizin Berlin.
Comprehensive evaluations show that Atlas achieves state-of-the-art performance across twenty-one public benchmark datasets, even though it is neither the largest model by parameter count nor by training dataset size.
\end{abstract}

\section{Introduction}
Anatomical pathology plays a central role in clinical medicine for tissue-based diagnostics and in biomedical research as a basis for the understanding of mechanisms of disease. Although molecular and omics-based data complement histological assessments, the microscopic evaluation of morphological changes remains the cornerstone of pathology practice. Consequently, with the advent of routine slide digitization, computational pathology has focused on making the analysis of histopathology slide images more precise, scalable, and reliable.

However, despite artificial intelligence (AI) having led to promising proof-of-concepts and applications (e.g., \cite{klauschen2024toward,diao2021human,raciti2023clinical,binder2021morphological,keyl2022patient}), generalization, application variety, and robustness remain challenging and have hindered the broad translation of AI applications into clinical routine diagnostics. Here, the limited adoption of digitization in clinical practice results in a scarcity of training data, particularly for infrequent and rare diseases~\cite{dippel2024ai}. Furthermore, generating sufficient labeled data representing the full spectrum of human disease, biological, and technical variability inherent to morphology, tissue processing, staining, and slide scanners has proven logistically and financially challenging.

Addressing these problems, foundation models have quickly gained traction in the pathology domain based on their promise to achieve robust and generalizable data representations by incorporating the diversity found in pathology through large-scale self-supervised training. The generalizability and robustness of foundation models are particularly relevant to the performance of downstream tasks in pathology---a domain that has both high variation in input data and tasks, such as disease diagnoses, outcome prediction, detection of morphological structures, and quantification of biomarkers.

\begin{figure}[t]
    \centering
    \includegraphics[width=\linewidth]{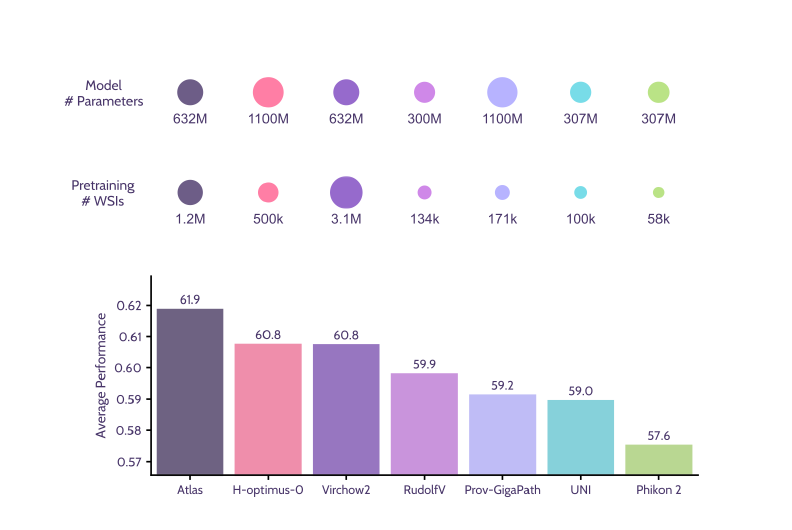}
    \caption{\small Overview of average performance, training dataset size, and model size of different contenders. The average performance is detailed in Table~\ref{tab:MAX}. H-Optimus-0~\cite{hoptimus0} and Prov-GigaPath~\cite{xu_whole-slide_2024} are the models with the most parameters and Virchow2~\cite{zimmermann_virchow_2024} is the model trained on most slides. Our model Atlas exhibits the best average performance and an intermediate model and training dataset size.}
    \label{fig:overview}
\end{figure}

In this report, we utilized a corpus of 1.2 million histopathology whole slide images derived from more than 490,000 cases from Mayo Clinic and Charité - Universitätsmedizin Berlin to train a ViT-H/14 pathology foundation model called ``Atlas'' using the training paradigm from RudolfV~\cite{dippel_rudolfv_2024}. Atlas incorporates a broad diversity of diseases, staining types, and scanners, and utilizes multiple image magnifications during training.  We provide an assessment of its performance compared to other leading models available for testing using twenty-one benchmark datasets assessing a variety of downstream pathology tasks. An overview of model characteristics and results can be found in Figure \ref{fig:overview}.

\section{Related Work}
Multiple previous works on histopathology foundation models have developed a variety of models with increasingly large parameter counts and on increasingly expansive pathology datasets curated from public and private sources \cite{dippel_rudolfv_2024,zimmermann_virchow_2024,filiot_phikon-v2_2024,xu_whole-slide_2024,hoptimus0,chen_towards_2024,ding2024multimodalslidefoundationmodel,lenz2024unsupervisedfoundationmodelagnosticslidelevel,juyal_pluto_2024,lu2024avisionlanguage,qiu2024pixelsgigapixelsbridginglocal}. Pathology foundation models can be separated into tile- and slide-based models. Tile-based models, which represent the majority of the models, generate embeddings from fixed-sized image tiles derived from gigapixel whole slide images (WSIs). Tile-based models have typically been trained via self-supervised-learning \cite{dippel_rudolfv_2024,zimmermann_virchow_2024,filiot_phikon-v2_2024,xu_whole-slide_2024,hoptimus0,chen_towards_2024,ai_towards_2024,nechaev_hibou_2024,wang_transformer-based_2022,kang2022benchmarking,lu2024avisionlanguage} and most studies \cite{dippel_rudolfv_2024,zimmermann_virchow_2024,filiot_phikon-v2_2024,xu_whole-slide_2024,hoptimus0,chen_towards_2024,ai_towards_2024,nechaev_hibou_2024} are based on the DINOv2 framework \cite{oquab2023dinov2}. We include the leading and available tile-based models in our study, i.e. Virchow2 (632 million parameters; 3.1 million slides; \cite{zimmermann_virchow_2024}), H-Optimus (1.1 billion parameters; 500k slides; \cite{hoptimus0}), RudolfV (300 million parameters; 134k slides; \cite{dippel_rudolfv_2024}), UNI (300 million parameters; 100k slides; \cite{chen_towards_2024}), and Phikon 2 (300 million parameters; 58k slides; \cite{filiot_phikon-v2_2024}).

Slide-based foundation models seek to create global representations of whole slide images, which requires encoding features that may span individual tiles or otherwise represent global phenomena not readily evident on single tiles. To address this, most current slide-based models operate in two steps~\cite{ding2024multimodalslidefoundationmodel,lenz2024unsupervisedfoundationmodelagnosticslidelevel,wang_pathology_2024,xu_whole-slide_2024,shaikovski_prism_2024}. First, by encoding individual tiles and secondly by aggregating the resulting tile representations into a slide-level representation. For such models, the tile-encoder is a key performance gateway. Accordingly, this study compares the tile-based encoder of Prov-GigaPath (1.1 billion parameters; 171k slides; \cite{xu_whole-slide_2024}).

To date, many foundation models in pathology have been trained exclusively on hematoxylin and eosin (H\&E) stained slides with tiles extracted at a single magnification level. While H\&E staining represents the bulk of routine pathology, alternative histochemical and immunohistochemical (IHC) stains are the basis for many biomarker evaluations and accordingly have been increasingly represented in recent studies \cite{zimmermann_virchow_2024,dippel_rudolfv_2024,xu_whole-slide_2024}. Similarly, data from multiple image magnifications have been included in recent work~\cite{zimmermann_virchow_2024,ai_towards_2024}. Atlas incorporates these features, and uses data obtained from H\&E, IHC, and special stains, as well as multiple magnifications.

\section{Data and Methods}

\subsection{Dataset and Preprocessing}
\begin{figure}[t]
    \centering
    \includegraphics[width=\linewidth]{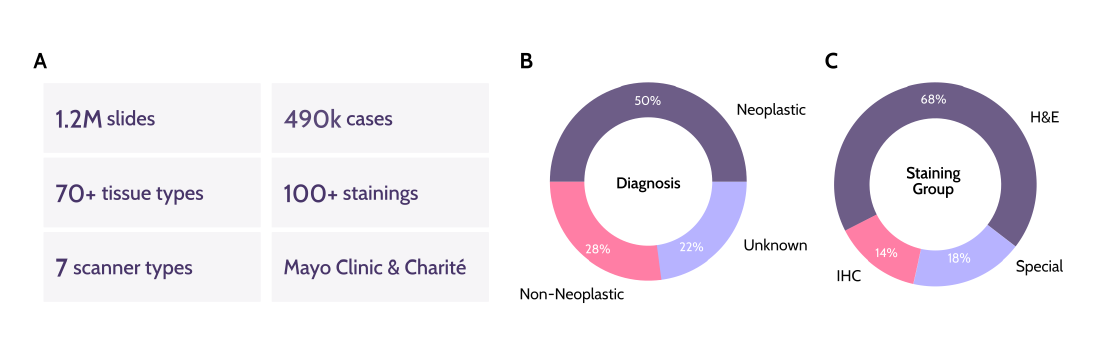}
    \caption{\small (A) shows the key training dataset statistics. The dataset was derived from 1.2 million pathology slides from 490k cases. The dataset contains data from over 70 tissue/organ types, over 100 different staining types, and 7 scanner types. The data was sourced from Mayo Clinic and Charité - Universitätsmedizin Berlin. (B) shows the distribution of neoplastic vs. non-neoplastic diseases. (C) shows the distribution of the staining groups H\&E, IHC, and special stains.}
    \label{fig:dataset}
\end{figure}

A curated set of 1.2 million de-identified WSIs, derived from 490k pathology cases, from the digital archives of Mayo Clinic and Charité - Universitätsmedizin Berlin, was utilized to generate 3.4 billion image tiles for training. Tiles were extracted at multiple resolutions, namely 0.25, 0.5, 1.0, and 2.0 microns per pixel, corresponding to objective microscopic magnifications of $40\times$, $20\times$, $10\times$, and $5\times$, respectively. An overview of relevant dataset statistics are given in Figure~\ref{fig:dataset}. Slides and tiles were sampled for training using the sampling algorithm of RudolfV~\cite{dippel_rudolfv_2024}.

\subsection{Model Framework and Compute Environment}
A sampled dataset of ca. 520m tiles was used to train a ViT-H/14 (632 million parameters) model~\cite{dosovitskiy2020vit} using an adapted RudolfV~\cite{dippel_rudolfv_2024} approach, which is based on the DinoV2 framework~\cite{oquab2023dinov2}. Model training was performed with Nvidia H100 GPUs within the Mayo Clinic Platform environment\footnote{https://www.mayoclinicplatform.org/}.

\subsection{Evaluation Protocols}
\label{ssec:evaluation_protocols}
We evaluated model performance using linear probing protocols as established in the literature \cite{dippel_rudolfv_2024,hoptimus0,zimmermann_virchow_2024}, with all models evaluated on extracted embeddings from frozen backbones. We use both public benchmarks as well as public evaluation frameworks where available, to foster reproducibility and comparability. Results were computed for 5 seeds over data split, shuffling, and ``probing'' model initialization per foundation model and task, if not stated otherwise in a specific task description (see Appendix~\ref{ssec:task_descriptions}). We report the mean performance and standard errors over the seeds. Dataset splits are described in detail in the respective task descriptions in Appendix~\ref{ssec:task_descriptions}. No augmentations were applied when extracting embeddings. As all models use Vision Transformer architectures \cite{dosovitskiy2020vit}, we always evaluate both the CLS token and the CLS+Mean\footnote{CLS+Mean being the concatenation of the CLS token and the mean over all ``mini-patch'' tokens} token embeddings for every model and report the better (maximum) performance of the two in Table~\ref{tab:MAX}. This accounts for potential systematic differences between information encoded in different tokens between different models, as also done in previous works (see e.g.\ \cite{zimmermann_virchow_2024}). Results of using the CLS and CLS+Mean token only, are reported in Table~\ref{tab:CLS} and Table~\ref{tab:cls_mean}, respectively.

\paragraph{Patch-level classification} For patch-level classification tasks, which make up the majority of benchmark datasets, linear probing (LP) is the default protocol. Balanced accuracy was utilized as the performance metric for all patch-level classification tasks.

We use \textit{eva} \cite{kaiko.ai2024eva}, an open-source evaluation framework for pathology foundation models, for LP evaluation where available\footnote{Available at \url{https://github.com/kaiko-ai/eva}, accessed on Oct 31, 2024 (version: 0.1.3)}. This includes the BACH~\cite{aresta2019bach}, CRC-100k~\cite{crc-100k}, MHIST~\cite{mhist}, and PCAM~\cite{pcam} datasets. Linear classification on extracted embeddings in \textit{eva} is done by training a single-layer neural network with a batch size of 256 patches using stochastic gradient descent (SGD) with a cosine learning rate schedule and base learning rate of 0.0003 for a total of 12.5k iterations. 

For patch-level classification tasks not implemented in \textit{eva}, we use an internal LP evaluation framework. These include the MSI CRC (patch), MSI STAD (patch), Pan-cancer TIL, TCGA Uniform (10x), and TCGA Uniform (20x) datasets. In the internal framework, we perform LP on extracted embeddings using \texttt{scikit-learn}'s Logistic Regression~\cite{scikit-learn} with balanced class weights. The L2 regularization parameter is chosen by performing a cross-validated grid search over 15 different values between $10^{-8}$ and $10^{4}$. Using the best parameter, a final model is fit and applied to the test set. 

\paragraph{Slide-level classification} For all slide-level classification tasks, we use the \textit{eva}~\cite{kaiko.ai2024eva} framework, which applies the default Attention-based Multiple Instance Learning (ABMIL) \cite{ilse2018attention} protocol. Here, an ABMIL head with ReLU activations is trained with a batch size of 32 slides using the AdamW optimizer \cite{loshchilov2018decoupled} with a cosine learning rate schedule and base learning rate of 0.001 for a total of 12.5k iterations. Balanced accuracy was utilized as a performance metric for all slide-level classification tasks.

\paragraph{Patch-level regression} The HEST-Benchmark \cite{jaume2024hest} is composed of tasks for gene expression prediction and designed as multivariate regression. We follow and use the default evaluation protocol as recommended and implemented\footnote{Available at \url{https://github.com/mahmoodlab/HEST}, accessed git commit '5a0cbba'.} by the authors. We provide respective details in the description of the HEST-Benchmark in Appendix~\ref{ssec:task_descriptions}.

\begin{table}[t]

\caption{\small Overview of results on different benchmark datasets. The benchmark datasets are split into morphology- and molecular-related benchmarks. The metric for the first 10 benchmark datasets is Pearson correlation, and the metric for the other benchmark datasets is balanced accuracy. Higher values are better, the highest value per row is bold, and the second-highest value is underlined. The evaluation protocols and descriptions for each benchmark dataset can be found in the methods~\ref{ssec:evaluation_protocols} and appendix~\ref{ssec:task_descriptions} sections, respectively.
\vspace{3mm}
} 
\label{tab:MAX}
\centering
\begin{adjustbox}{width=\textwidth}
\small

\begin{tabular}{p{1cm}@{}wl{2cm}wr{1.9cm}wr{1.9cm}wr{1.9cm}wr{1.9cm}wr{1.9cm}wr{1.9cm}wr{1.9cm}} 
Group & Benchmark & Phikon v2~\cite{filiot_phikon-v2_2024} & UNI~\cite{chen_towards_2024} & Gigapath~\cite{xu_whole-slide_2024} & RudolfV~\cite{dippel_rudolfv_2024}  & Virchow2~\cite{zimmermann_virchow_2024} & H-optimus-0~\cite{hoptimus0} & Atlas \\ 
\midrule 
\parbox[t]{2mm}{\multirow{12}{*}{\rotatebox[origin=c]{90}{Molecular-related}}} & HEST-COAD & 25.6 & 26.2 & 30.7 & \textbf{31.0} & 25.9 & \underline{30.9} & 29.4 \\ 
  & HEST-HCC & 7.8 & 8.3 & 7.1 & 9.4 & \underline{9.6} & 8.4 & \textbf{10.7} \\ 
  & HEST-IDC & 56.6 & 58.5 & 56.8 & 57.4 & 59.3 & \textbf{61.0} & \underline{60.4} \\ 
  & HEST-LUAD & 54.8 & 55.2 & 55.8 & \underline{57.7} & 56.9 & 57.3 & \textbf{58.0} \\ 
  & HEST-LYMPH\_IDC & 24.8 & 25.8 & 25.1 & 25.6 & 25.9 & \textbf{26.8} & \underline{26.4} \\ 
  & HEST-PAAD & 47.9 & 48.8 & 49.5 & \underline{51.1} & 47.3 & 50.9 & \textbf{51.8} \\ 
  & HEST-PRAD & 37.7 & 32.2 & \underline{38.4} & 37.7 & 35.1 & \textbf{38.5} & 38.4 \\ 
  & HEST-READ & 18.5 & 18.4 & 19.6 & 19.9 & 21.1 & \textbf{24.1} & \underline{22.8} \\ 
  & HEST-SKCM & 58.4 & 63.5 & 58.8 & 61.8 & \underline{63.7} & \textbf{66.1} & 62.5 \\ 
  & HEST-ccRCC & 27.3 & 25.3 & 24.9 & 25.3 & 27.4 & \underline{29.0} & \textbf{29.4} \\ 
  & MSI CRC (patch) & 68.8 & 69.5 & 70.4 & 69.9 & \textbf{74.0} & 71.2 & \underline{73.6} \\ 
  & MSI STAD (patch) & 71.2 & 70.5 & 71.0 & 74.1 & \underline{74.8} & 73.6 & \textbf{76.0} \\ 
 [0.05cm]\multicolumn{2}{c}{Molecular-Average} & 41.6 & 41.8 & 42.3 & 43.4 & 43.4 & \underline{44.8} & \textbf{44.9} \\ 
 \midrule 
\parbox[t]{2mm}{\multirow{9}{*}{\rotatebox[origin=c]{90}{Morphology-related}}} & Pan-cancer TIL & 92.9 & 92.6 & 92.3 & 92.6 & \textbf{93.1} & \underline{93.0} & 93.0 \\ 
  & TCGA Uniform (10x) & 64.0 & 68.6 & 69.1 & 70.6 & \textbf{73.0} & 70.4 & \underline{71.8} \\ 
  & TCGA Uniform (20x) & 69.8 & 67.8 & 68.0 & \textbf{78.1} & 71.5 & \underline{72.4} & 67.8 \\ 
  & BACH & 73.8 & 80.1 & 80.2 & 76.9 & \underline{88.7} & 75.8 & \textbf{93.1} \\ 
  & CRC-100k & 95.5 & 95.4 & 95.9 & 96.0 & \underline{96.7} & 96.2 & \textbf{97.1} \\ 
  & MHIST & 78.4 & 84.4 & 83.1 & 80.5 & \underline{85.9} & 85.0 & \textbf{86.4} \\ 
  & PCAM & 90.0 & 93.6 & 94.5 & \underline{94.6} & 93.9 & 94.3 & \textbf{94.9} \\ 
  & CAMELYON16 & 79.8 & 85.0 & 82.1 & 77.1 & \underline{86.5} & 84.0 & \textbf{86.8} \\ 
  & PANDA & 65.3 & \underline{69.6} & 69.6 & \underline{69.6} & 66.4 & 68.0 & \textbf{70.5} \\ 
 [0.05cm]\multicolumn{2}{c}{Morphology-Average} & 78.8 & 81.9 & 81.6 & 81.8 & \underline{84.0} & 82.1 & \textbf{84.6} \\ 
 \midrule 
\multicolumn{2}{c}{Overall Average} & 57.6 & 59.0 & 59.2 & 59.9 & 60.8 & \underline{60.8} & \textbf{61.9} \\ 
 \bottomrule 
\end{tabular}

\end{adjustbox}
\end{table} 

\section{Results}
The following analysis is based on 21 public benchmark datasets from two public foundation model evaluation frameworks \textit{eva}~\cite{kaiko.ai2024eva} and HEST~\cite{jaume2024hest} as well as additional tumor-micro-environment (TME) and cancer typing benchmark datasets. The tasks range from TME tissue- and cell-typing over identifying morphological patterns, identifying cancer subtypes, to classifying molecular mutations. The task descriptions and evaluation protocols are detailed in Appendix~\ref{ssec:task_descriptions} and Section~\ref{ssec:evaluation_protocols}, respectively.

Atlas achieves an average performance score of 61.9\%, a 1.1 p.p. improvement over the two closest contenders, Virchow2~\cite{zimmermann_virchow_2024} and H-Optimus-0~\cite{hoptimus0} (see Table~\ref{tab:MAX}). It displayed the highest performance on 6 out of 9 morphology-related tasks with Virchow2 being the closest contender performing best on 2 tasks. For the molecular-related biomarker tasks H-Optimus and Atlas perform both best on 5 out of 12 tasks.  Overall, Atlas displayed the best performance on 11 of 21 assessed tasks across both molecular- and morphology-related tasks. In 7 of the 10 benchmarks in which Atlas was not the leading model it was the second best model by performance. The model displayed below-average performance for a single benchmark, TCGA Uniform (20$\times$), where Atlas and UNI were the poorest performers; interestingly, Atlas' performance on the 10$\times$ version of that benchmark was second only to Virchow2. 

\section{Discussion}
Our novel pathology foundation model Atlas showed consistently good results across a diverse set of benchmarks covering molecular and morphology related tasks. Related work using a larger data base~\cite{zimmermann_virchow_2024} or more model parameters \cite{zimmermann_virchow_2024,hoptimus0,xu_whole-slide_2024} suggests that scaling data and model size even further might yield additional improvements.

To accurately assess the quality, generalization, and robustness of pathology foundation models, it is essential to utilize a wide range of (public) benchmark datasets, covering the entire spectrum of diseases, morphologies, scanners, etc. Despite first standardized frameworks such as \textit{eva}~\cite{kaiko.ai2024eva} and HEST~\cite{jaume2024hest} being available, foundation model development would benefit from a larger and more diverse pool of benchmark datasets. Access to such diverse datasets would provide deeper insights into model performance and robustness, thereby enhancing our understanding of the capabilities and limitations of pathology foundation models.

\section*{Acknowledgments}
We would like to thank the teams at Aignostics, Charité - Universitätsmedizin Berlin - Pathology Department, Mayo Clinical Platform, Mayo Clinic Generative Artificial Intelligence Program, Mayo Clinic Department of Laboratory Medicine and Pathology, and Mayo Clinic Digital Pathology for supporting this work. 
Special thanks goes to Alexander Baxendale, Amelie Froessl, Amin Abbasloo, Barbara Feulner, Carl Andersson, Dharma Indurthy, Erinç Argımak, Fabian Spie{\ss}, Gerrit Erdmann, Jay Tolley, Jeannine Korp, Jeff Anderson, Jennifer Flores, Joshua Pankratz, Julika Ribbat-Idel, Mark Ibrahim, Olja Smilić, Rob Blundo, Tom Lehmann, Sara Then, Steele Clifton-Berry, and Valentin François for their organizational, technical, and other support.

The benchmark results shown here are in part based upon data generated by the TCGA Research Network: https://www.cancer.gov/tcga.

This work was in part supported by the German Ministry for Education and Research (BMBF) under Grants 01IS14013A-E, 01GQ1115, 01GQ0850, 01IS18025A,
031L0207D, and 01IS18037A.  K.R.M. was partly supported by the Institute of Information \& Communications Technology Planning \& Evaluation (IITP) grants funded by the Korea government (MSIT) (No. 2019-0-00079, Artificial Intelligence Graduate School Program, Korea University and No. 2022-0-00984, Development of Artificial Intelligence Technology for Personalized Plug-and-Play Explanation and Verification of Explanation).


\bibliographystyle{plain} 
\bibliography{references} 

\newpage
\appendix

\section{Appendix}

\subsection{Task Descriptions}
\label{ssec:task_descriptions}

\paragraph{HEST-Benchmark} The HEST-Benchmark was introduced by \cite{jaume2024hest} for benchmarking foundation models for histology on the task of gene expression prediction from H\&E-stained images. The benchmark includes 72 spatial transcriptomics profiles (using Xenium or Visium technology) with corresponding H\&E-stained images from 47 patients grouped into 10 tasks based on organ. Each task involves predicting the expression levels of the 50 most variable genes (highest normalized variance) from 112$\times$112 {\textmu}m H\&E-stained image patches (equivalent to 224$\times$224 pixels at 20$\times$ magnification) centered on each spatial transcriptomics spot, formulated as a multivariate regression problem. We used the default Ridge Regression with PCA (256 factors) evaluation protocol to solve the multivariate regression on extracted embeddings \cite{jaume2024hest}. Specifically, the 10 tasks are to predict gene expression levels for invasive ductal carcinoma (breast cancer, IDC), prostate adenocarcinoma (prostate cancer, PRAD), pancreatic adenocarcinoma (pancreatic cancer, PAAD), skin cutaneous melanoma (skin cancer, SKCM), colonic adenocarcinoma (colon cancer, COAD), rectal adenocarcinoma (rectum cancer, READ), clear cell renal cell carcinoma (kidney cancer, ccRCC), hepatocellular carcinoma (liver cancer, HCC), lung adenocarcinoma (lung cancer, LUAD), and axillary lymph nodes in IDC (metastatic, LYMPH-IDC). The benchmark applies patient-stratified splits to avoid any train/test patient-level data leakage, resulting in a k-fold cross-validation where k is the number of patients \cite{jaume2024hest}. Performance is evaluated using the Pearson correlation between the predicted and measured gene expression and reported results are the mean and standard deviation across folds.

\paragraph{MSI CRC (patch)} This dataset contains 173,630 H\&E images (224$\times$224 pixels at 20$\times$ magnification) extracted from $N =$ 360 colorectal cancer (CRC) tissue scans from TCGA (TCGA-CRC-DX). The task is binary classification of microsatellite instability (MSI) vs.\ microsatellite stability (MSS), which is a clinically important prognostic marker \cite{kather2019deep,kaczmarzyk2023champkit}. The dataset is split into 56,044 (28,022 MSI + 28,022 MSS) training images, 18,682 (9,341 MSI + 9,341 MSS) validation images, and 98,904 (28,335 MSI + 70,569 MSS) test images.

\paragraph{MSI STAD (patch)} This dataset contains 198,464 H\&E images (224$\times$224 pixels at 20$\times$ magnification) extracted from $N =$ 315 stomach adenocarcinoma (STAD) tissue scans from TCGA (TCGA-STAD). The task is binary classification of microsatellite instability (MSI) vs.\ microsatellite stability (MSS), which is a clinically important prognostic marker \cite{kather2019deep,kaczmarzyk2023champkit}. The dataset is split into 60,342 (30,171 MSI + 30,171 MSS) training images, 20,114 (10,057 MSI + 10,057 MSS) validation images, and 118,008 (27,904 MSI + 90,104 MSS) test images.

\paragraph{Pan-cancer TIL} The pan-cancer tumor-infiltrating lymphocytes (TIL) detection dataset contains 304,097 H\&E images (100$\times$100 pixels at 20$\times$ magnification) extracted from tissue sample scans of 23 different cancer types from TCGA \cite{abousamra2022deep,saltz2018spatial}. The task is to classify an image into TIL positive vs.\ negative. An image is labeled positive if it contains at least two TILs. The dataset is split into 209,221 training images and 56,275 test images.

\paragraph{TCGA Uniform (10$\times$) and (20$\times$)} The TCGA Uniform dataset \cite{komura_tcga_uniform} contains 264,110 to 271,700 patches per resolution with $256 \times 256$ pixels. The task is to differentiate between 32 cancer classes from different tissue types (e.g., Colon adenocarcinoma). Only patches showing the specific cancer type were extracted from the TCGA WSIs based on annotations from two trained pathologists.
As there is no official train and test split, we generated five folds with no overlapping patients and sampled 100 patches per class and fold, resulting in a total dataset size of 16.000 patches. We generated one dataset containing patches with \SI{0.5}{\micro\metre}/pixel (20$\times$) and one with \SI{1.0}{\micro\metre}/pixel (10$\times$) to test the performance of the foundation models at different zoom levels. The results represent the mean balanced accuracy on the five-fold cross-validation evaluation.


\paragraph{BACH} The BACH dataset comprises 400 H\&E microscopy images (2048$\times$1536 pixels at 20$\times$ magnification) of breast cancer biopsies. It originates from the ICIAR 2018 Grand Challenge on BreAst Cancer Histology images \cite{aresta2019bach}. The classification task of the challenge is to classify each image into one of the following four classes: normal, benign, in situ carcinoma, and invasive carcinoma. The dataset is split into 268 training images (67 images per class) and 132 test images (33 images per class). The patches are resized and cropped to 224$\times$224 pixels.

\paragraph{CRC-100k} The CRC-100k dataset contains 107,180 H\&E images (224$\times$224 pixels at 20$\times$ magnification) extracted from colorectal cancer (CRC) tissue samples \cite{kather2019predicting}. The tissue samples originate from CRC primary tumors and CRC liver metastases. The task of this benchmark is to classify each image into one of the following 9 tissue classes: adipose, background, debris, lymphocytes, mucus, smooth muscle, normal colon mucosa, cancer-associated stroma, and colorectal adenocarcinoma epithelium. The dataset is split into 100,000 training images (NCT-CRC-HE-100K-NONORM) and 7,180 test images (CRC-VAL-HE-7K). We take the original (no-norm) images without Macenko color normalization~\cite{crc-100k}.

\paragraph{MHIST} The task of MHIST is to classify images of colorectal polyps into hyperplastic polyps (HPs) vs.\ sessile serrated adenomas (SSAs) \cite{mhist}. This distinction is clinically important as HPs are typically benign whereas SSAs are precancerous lesions that can turn into cancer if left untreated. The task is challenging for pathologists, often showing considerable inter-pathologist variability. The MHIST dataset consists of 3,152 H\&E images (224$\times$224 pixels at 8$\times$ magnification) of colorectal polyps and labels are derived from the majority vote of seven pathologists. The dataset is split into 2,162 training images (1,545 HP and 617 SSA) and 990 test images (630 HP and 360 SSA). 

\paragraph{PCAM} PCAM (PatchCamelyon) defines the clinically-relevant task of metastasis detection as a binary image classification task \cite{pcam}. The dataset consists of 327,680 H\&E images (96$\times$96 pixels at 10$\times$ magnification) extracted from scans of sentinel lymph node sections. Each image is annotated with a binary label indicating the presence of metastatic breast cancer tissue. An image is labeled as metastatic if the center 32$\times$32 pixels region contains at least one pixel of tumor tissue. The dataset is split by 80:10:10 into training, validation, and test sets with no overlap of WSIs/cases between the splits and every split having a 50:50 balance of positive and negative examples. For evaluation, we resize each image to 224x224 pixels. PCam has been derived from the CAMELYON16 Challenge \cite{camelyon}.

\paragraph{CAMELYON16} The task of the CAMELYON16 challenge \cite{camelyon} is to classify whole slide images (WSIs) of lymph node tissue sections into having metastatic breast cancer tissue or not. The dataset comprises 399 H\&E-stained WSIs of sentinel lymph node sections, which were acquired and scanned (40$\times$ magnification) at two centers from the Netherlands \cite{camelyon}. The dataset is split into 270 training slides (110 with and 160 without metastasis) and 129 test slides (49 with and 80 without metastasis). Here, we report results for the CAMELYON16 (small) setup in \textit{eva} \cite{kaiko.ai2024eva}, which randomly samples max.\ 1,000 patches per slide.

\paragraph{PANDA} The PANDA challenge \cite{panda} considers the challenging task of tumor grading of whole slide images (WSIs) of prostate cancer biopsies, which suffers from significant inter-observer variability between pathologists. Prostate cancer grading follows the Gleason grading system (3, 4, or 5) based on architectural growth patterns of the tumor, which are then converted into an ISUP grade on a scale of 1-5 for use as a prognostic marker. The dataset features 9,555 H\&E-stained WSIs (subset with noisy labels removed) of prostate tissue biopsies from two medical centers scanned at 20$\times$ magnification. Specifically, the task is to classify each WSI into an ISUP grade of 0--5 (six classes), where 0 means that a biopsy does not contain any cancer. The dataset is split into 6,686 training slides, 1,430 validation slides, and 1,439 test slides in a class-stratified manner. Here, we report results for the PANDA (small) setup in \textit{eva} \cite{kaiko.ai2024eva}, which considers a fewer number of total slides (955 train, 477 validation, 477 test) as well as fewer randomly sampled patches per slide (200).

\begin{table}[h]
\caption{Summary of benchmark datasets and evaluation frameworks.
\vspace{3mm}} 
\label{tab:DS_INF}
\centering
\begin{adjustbox}{width=1\textwidth}
\scriptsize
\begin{tabular}{llllrl}
Dataset & Pathological Task & ML Task & Input Type & Size & Implementation\\
\noalign{\vspace{3pt}}
\midrule
\noalign{\vspace{3pt}}
HEST-Benchmark & Gene Expression Prediction & Regression & Patch & 72 & HEST \\
MSI CRC (patch) & Microsatellite Instability Prediction & Classification (binary) & Patch & 173,630 & Internal \\
MSI STAD (patch) & Microsatellite Instability Prediction & Classification (binary) & Patch & 198,464 & Internal \\
Pan-cancer TIL & Tumor-Infiltrating Lymphocytes Detection & Classification (binary) & Patch & 304,097 & Internal  \\
TCGA Uniform (10$\times$) & Cancer Subtyping & Classification & Patch & 264,110 & Internal \\
TCGA Uniform (20$\times$) & Cancer Subtyping & Classification & Patch & 271,700 & Internal \\
BACH & Breast Cancer Classification & Classification & Patch & 400 & \textit{eva} \\
MHIST & Colorectal Polyp Classification & Classification & Patch & 3,152 & \textit{eva} \\
PCAM & Metastasis Detection  & Classification & Patch & 327,680 & \textit{eva} \\
CRC-100k & Tissue Classification & Classification & Patch & 107,180 & \textit{eva} \\
CAMELYON16 & Metastasis Detection  & Classification & WSI & 399 & \textit{eva} \\
PANDA (small) &  Tumor Grading & Classification & WSI & 1909 & \textit{eva} \\
\bottomrule
\end{tabular}
\end{adjustbox}
\end{table} 

\FloatBarrier

\subsection{Additional Results}




\begin{table}[ht]
\caption{Model results per task, maximum over CLS token and CLS+MEAN token. Same table as Table~\ref{tab:MAX} but with standard deviation over the 5 data splits.
\vspace{3mm}
} 
\label{tab:MAX_std}
\centering
\begin{adjustbox}{width=1\textwidth}
\small

\begin{tabular}{p{1cm}@{}wl{2cm}wr{1.9cm}wr{1.9cm}wr{1.9cm}wr{1.9cm}wr{1.9cm}wr{1.9cm}wr{1.9cm}} 
Group & Benchmark & Phikon v2~\cite{filiot_phikon-v2_2024} & UNI~\cite{chen_towards_2024} & Gigapath~\cite{xu_whole-slide_2024} & RudolfV~\cite{dippel_rudolfv_2024}  & Virchow2~\cite{zimmermann_virchow_2024} & H-optimus-0~\cite{hoptimus0} & Atlas \\ 
\midrule 
\parbox[t]{2mm}{\multirow{12}{*}{\rotatebox[origin=c]{90}{Molecular-related}}} & HEST-COAD & 25.6 \scriptsize{$\pm 2.3$}  & 26.2 \scriptsize{$\pm 3.1$}  & 30.7 \scriptsize{$\pm 0.0$}  & \textbf{31.0} \scriptsize{$\pm 2.0$}  & 25.9 \scriptsize{$\pm 1.6$}  & \underline{30.9} \scriptsize{$\pm 0.0$}  & 29.4 \scriptsize{$\pm 1.5$}  \\ 
  & HEST-HCC & 7.8 \scriptsize{$\pm 1.2$}  & 8.3 \scriptsize{$\pm 0.5$}  & 7.1 \scriptsize{$\pm 1.3$}  & 9.4 \scriptsize{$\pm 1.7$}  & \underline{9.6} \scriptsize{$\pm 1.0$}  & 8.4 \scriptsize{$\pm 1.2$}  & \textbf{10.7} \scriptsize{$\pm 1.9$}  \\ 
  & HEST-IDC & 56.6 \scriptsize{$\pm 7.8$}  & 58.5 \scriptsize{$\pm 7.7$}  & 56.8 \scriptsize{$\pm 7.6$}  & 57.4 \scriptsize{$\pm 8.5$}  & 59.3 \scriptsize{$\pm 8.5$}  & \textbf{61.0} \scriptsize{$\pm 8.1$}  & \underline{60.4} \scriptsize{$\pm 8.3$}  \\ 
  & HEST-LUAD & 54.8 \scriptsize{$\pm 2.3$}  & 55.2 \scriptsize{$\pm 2.2$}  & 55.8 \scriptsize{$\pm 2.9$}  & \underline{57.7} \scriptsize{$\pm 1.8$}  & 56.9 \scriptsize{$\pm 1.7$}  & 57.3 \scriptsize{$\pm 2.7$}  & \textbf{58.0} \scriptsize{$\pm 1.5$}  \\ 
  & HEST-LYMPH\_IDC & 24.8 \scriptsize{$\pm 4.9$}  & 25.8 \scriptsize{$\pm 4.1$}  & 25.1 \scriptsize{$\pm 4.2$}  & 25.6 \scriptsize{$\pm 3.3$}  & 25.9 \scriptsize{$\pm 3.3$}  & \textbf{26.8} \scriptsize{$\pm 4.0$}  & \underline{26.4} \scriptsize{$\pm 4.4$}  \\ 
  & HEST-PAAD & 47.9 \scriptsize{$\pm 7.0$}  & 48.8 \scriptsize{$\pm 5.8$}  & 49.5 \scriptsize{$\pm 5.8$}  & \underline{51.1} \scriptsize{$\pm 7.8$}  & 47.3 \scriptsize{$\pm 6.9$}  & 50.9 \scriptsize{$\pm 4.3$}  & \textbf{51.8} \scriptsize{$\pm 7.4$}  \\ 
  & HEST-PRAD & 37.7 \scriptsize{$\pm 0.1$}  & 32.2 \scriptsize{$\pm 8.1$}  & \underline{38.4} \scriptsize{$\pm 3.1$}  & 37.7 \scriptsize{$\pm 2.6$}  & 35.1 \scriptsize{$\pm 3.4$}  & \textbf{38.5} \scriptsize{$\pm 0.0$}  & 38.4 \scriptsize{$\pm 0.7$}  \\ 
  & HEST-READ & 18.5 \scriptsize{$\pm 5.9$}  & 18.4 \scriptsize{$\pm 4.9$}  & 19.6 \scriptsize{$\pm 6.2$}  & 19.9 \scriptsize{$\pm 6.8$}  & 21.1 \scriptsize{$\pm 5.0$}  & \textbf{24.1} \scriptsize{$\pm 2.7$}  & \underline{22.8} \scriptsize{$\pm 3.1$}  \\ 
  & HEST-SKCM & 58.4 \scriptsize{$\pm 6.2$}  & 63.5 \scriptsize{$\pm 3.6$}  & 58.8 \scriptsize{$\pm 5.8$}  & 61.8 \scriptsize{$\pm 4.6$}  & \underline{63.7} \scriptsize{$\pm 3.1$}  & \textbf{66.1} \scriptsize{$\pm 5.8$}  & 62.5 \scriptsize{$\pm 2.4$}  \\ 
  & HEST-ccRCC & 27.3 \scriptsize{$\pm 4.0$}  & 25.3 \scriptsize{$\pm 3.8$}  & 24.9 \scriptsize{$\pm 4.1$}  & 25.3 \scriptsize{$\pm 5.2$}  & 27.4 \scriptsize{$\pm 4.5$}  & \underline{29.0} \scriptsize{$\pm 3.8$}  & \textbf{29.4} \scriptsize{$\pm 5.5$}  \\ 
  & MSI CRC (patch) & 68.8 \scriptsize{$\pm 0.1$}  & 69.5 \scriptsize{$\pm 0.0$}  & 70.4 \scriptsize{$\pm 0.1$}  & 69.9 \scriptsize{$\pm 0.1$}  & \textbf{74.0} \scriptsize{$\pm 0.0$}  & 71.2 \scriptsize{$\pm 0.1$}  & \underline{73.6} \scriptsize{$\pm 0.0$}  \\ 
  & MSI STAD (patch) & 71.2 \scriptsize{$\pm 0.1$}  & 70.5 \scriptsize{$\pm 0.0$}  & 71.0 \scriptsize{$\pm 0.1$}  & 74.1 \scriptsize{$\pm 0.1$}  & \underline{74.8} \scriptsize{$\pm 0.2$}  & 73.6 \scriptsize{$\pm 0.0$}  & \textbf{76.0} \scriptsize{$\pm 0.1$}  \\ 
 [0.05cm]\multicolumn{2}{c}{Molecular-Average} & 41.6 & 41.8 & 42.3 & 43.4 & 43.4 & \underline{44.8} & \textbf{44.9} \\ 
 \midrule 
\parbox[t]{2mm}{\multirow{9}{*}{\rotatebox[origin=c]{90}{Morphology-related}}} & Pan-cancer TIL & 92.9 \scriptsize{$\pm 0.0$}  & 92.6 \scriptsize{$\pm 0.0$}  & 92.3 \scriptsize{$\pm 0.0$}  & 92.6 \scriptsize{$\pm 0.1$}  & \textbf{93.1} \scriptsize{$\pm 0.1$}  & \underline{93.0} \scriptsize{$\pm 0.1$}  & 93.0 \scriptsize{$\pm 0.0$}  \\ 
  & TCGA Uniform (10x) & 64.0 \scriptsize{$\pm 0.0$}  & 68.6 \scriptsize{$\pm 0.0$}  & 69.1 \scriptsize{$\pm 0.0$}  & 70.6 \scriptsize{$\pm 0.0$}  & \textbf{73.0} \scriptsize{$\pm 0.0$}  & 70.4 \scriptsize{$\pm 0.0$}  & \underline{71.8} \scriptsize{$\pm 0.0$}  \\ 
  & TCGA Uniform (20x) & 69.8 \scriptsize{$\pm 0.0$}  & 67.8 \scriptsize{$\pm 0.0$}  & 68.0 \scriptsize{$\pm 0.0$}  & \textbf{78.1} \scriptsize{$\pm 0.0$}  & 71.5 \scriptsize{$\pm 0.0$}  & \underline{72.4} \scriptsize{$\pm 0.0$}  & 67.8 \scriptsize{$\pm 0.0$}  \\ 
  & BACH & 73.8 \scriptsize{$\pm 0.3$}  & 80.1 \scriptsize{$\pm 1.0$}  & 80.2 \scriptsize{$\pm 0.3$}  & 76.9 \scriptsize{$\pm 0.3$}  & \underline{88.7} \scriptsize{$\pm 0.3$}  & 75.8 \scriptsize{$\pm 1.1$}  & \textbf{93.1} \scriptsize{$\pm 0.2$}  \\ 
  & CRC-100k & 95.5 \scriptsize{$\pm 0.0$}  & 95.4 \scriptsize{$\pm 0.2$}  & 95.9 \scriptsize{$\pm 0.0$}  & 96.0 \scriptsize{$\pm 0.1$}  & \underline{96.7} \scriptsize{$\pm 0.1$}  & 96.2 \scriptsize{$\pm 0.1$}  & \textbf{97.1} \scriptsize{$\pm 0.1$}  \\ 
  & MHIST & 78.4 \scriptsize{$\pm 0.4$}  & 84.4 \scriptsize{$\pm 0.1$}  & 83.1 \scriptsize{$\pm 0.1$}  & 80.5 \scriptsize{$\pm 0.0$}  & \underline{85.9} \scriptsize{$\pm 0.1$}  & 85.0 \scriptsize{$\pm 0.2$}  & \textbf{86.4} \scriptsize{$\pm 0.1$}  \\ 
  & PCAM & 90.0 \scriptsize{$\pm 0.1$}  & 93.6 \scriptsize{$\pm 0.1$}  & 94.5 \scriptsize{$\pm 0.0$}  & \underline{94.6} \scriptsize{$\pm 0.1$}  & 93.9 \scriptsize{$\pm 0.1$}  & 94.3 \scriptsize{$\pm 0.1$}  & \textbf{94.9} \scriptsize{$\pm 0.0$}  \\ 
  & CAMELYON16 & 79.8 \scriptsize{$\pm 1.4$}  & 85.0 \scriptsize{$\pm 0.5$}  & 82.1 \scriptsize{$\pm 0.7$}  & 77.1 \scriptsize{$\pm 1.7$}  & \underline{86.5} \scriptsize{$\pm 0.2$}  & 84.0 \scriptsize{$\pm 0.8$}  & \textbf{86.8} \scriptsize{$\pm 0.4$}  \\ 
  & PANDA & 65.3 \scriptsize{$\pm 0.4$}  & \underline{69.6} \scriptsize{$\pm 0.6$}  & 69.6 \scriptsize{$\pm 0.7$}  & \underline{69.6} \scriptsize{$\pm 0.4$}  & 66.4 \scriptsize{$\pm 1.1$}  & 68.0 \scriptsize{$\pm 0.6$}  & \textbf{70.5} \scriptsize{$\pm 0.5$}  \\ 
 [0.05cm]\multicolumn{2}{c}{Morphology-Average} & 78.8 & 81.9 & 81.6 & 81.8 & \underline{84.0} & 82.1 & \textbf{84.6} \\ 
 \midrule 
\multicolumn{2}{c}{Overall Average} & 57.6 & 59.0 & 59.2 & 59.9 & 60.8 & \underline{60.8} & \textbf{61.9} \\ 
 \bottomrule 
\end{tabular} 

\end{adjustbox}
\end{table}

\begin{table}[ht]
\caption{Model results per task, CLS token only.
\vspace{3mm}} 
\label{tab:CLS}
\centering
\begin{adjustbox}{width=1\textwidth}
\small

\begin{tabular}{p{1cm}@{}wl{2cm}wr{1.9cm}wr{1.9cm}wr{1.9cm}wr{1.9cm}wr{1.9cm}wr{1.9cm}wr{1.9cm}} 
Group & Benchmark & Phikon v2~\cite{filiot_phikon-v2_2024} & UNI~\cite{chen_towards_2024} & Gigapath~\cite{xu_whole-slide_2024} & RudolfV~\cite{dippel_rudolfv_2024}  & Virchow2~\cite{zimmermann_virchow_2024} & H-optimus-0~\cite{hoptimus0} & Atlas \\ 
\midrule 
\parbox[t]{2mm}{\multirow{12}{*}{\rotatebox[origin=c]{90}{Molecular-related}}} & HEST-COAD & 25.0 \scriptsize{$\pm 1.7$}  & 26.2 \scriptsize{$\pm 3.1$}  & \underline{29.9} \scriptsize{$\pm 2.1$}  & 23.7 \scriptsize{$\pm 6.0$}  & 25.9 \scriptsize{$\pm 1.6$}  & \textbf{30.9} \scriptsize{$\pm 0.0$}  & 25.9 \scriptsize{$\pm 3.1$}  \\ 
  & HEST-HCC & 6.7 \scriptsize{$\pm 1.3$}  & 7.8 \scriptsize{$\pm 0.2$}  & 7.1 \scriptsize{$\pm 1.3$}  & 6.5 \scriptsize{$\pm 0.1$}  & \underline{8.2} \scriptsize{$\pm 1.0$}  & 7.9 \scriptsize{$\pm 0.6$}  & \textbf{9.4} \scriptsize{$\pm 0.8$}  \\ 
  & HEST-IDC & 54.1 \scriptsize{$\pm 7.7$}  & 57.4 \scriptsize{$\pm 7.9$}  & 55.1 \scriptsize{$\pm 7.3$}  & 54.5 \scriptsize{$\pm 8.9$}  & 59.2 \scriptsize{$\pm 8.0$}  & \textbf{59.8} \scriptsize{$\pm 8.5$}  & \underline{59.6} \scriptsize{$\pm 8.1$}  \\ 
  & HEST-LUAD & 54.2 \scriptsize{$\pm 1.1$}  & 54.6 \scriptsize{$\pm 2.2$}  & 54.1 \scriptsize{$\pm 3.6$}  & 55.5 \scriptsize{$\pm 1.2$}  & 55.3 \scriptsize{$\pm 1.7$}  & \underline{55.9} \scriptsize{$\pm 3.3$}  & \textbf{57.0} \scriptsize{$\pm 1.7$}  \\ 
  & HEST-LYMPH\_IDC & 24.4 \scriptsize{$\pm 4.6$}  & 25.6 \scriptsize{$\pm 4.4$}  & 25.0 \scriptsize{$\pm 5.0$}  & 24.3 \scriptsize{$\pm 3.5$}  & 25.5 \scriptsize{$\pm 2.6$}  & \textbf{25.9} \scriptsize{$\pm 4.0$}  & \underline{25.7} \scriptsize{$\pm 4.7$}  \\ 
  & HEST-PAAD & 44.5 \scriptsize{$\pm 6.6$}  & 48.1 \scriptsize{$\pm 7.0$}  & 47.5 \scriptsize{$\pm 4.8$}  & 45.7 \scriptsize{$\pm 5.8$}  & 47.2 \scriptsize{$\pm 6.5$}  & \underline{49.1} \scriptsize{$\pm 4.0$}  & \textbf{50.7} \scriptsize{$\pm 7.2$}  \\ 
  & HEST-PRAD & 35.4 \scriptsize{$\pm 1.5$}  & 29.4 \scriptsize{$\pm 8.5$}  & \underline{37.0} \scriptsize{$\pm 2.1$}  & 37.0 \scriptsize{$\pm 2.8$}  & 34.8 \scriptsize{$\pm 3.1$}  & \textbf{38.5} \scriptsize{$\pm 0.0$}  & 35.3 \scriptsize{$\pm 3.2$}  \\ 
  & HEST-READ & 17.5 \scriptsize{$\pm 5.9$}  & 18.4 \scriptsize{$\pm 4.9$}  & 19.6 \scriptsize{$\pm 6.2$}  & 17.6 \scriptsize{$\pm 8.1$}  & 20.9 \scriptsize{$\pm 5.0$}  & \textbf{22.2} \scriptsize{$\pm 4.8$}  & \underline{21.3} \scriptsize{$\pm 2.9$}  \\ 
  & HEST-SKCM & 55.5 \scriptsize{$\pm 3.6$}  & \underline{63.5} \scriptsize{$\pm 3.6$}  & 56.1 \scriptsize{$\pm 6.2$}  & 58.0 \scriptsize{$\pm 4.1$}  & 61.9 \scriptsize{$\pm 2.8$}  & \textbf{64.5} \scriptsize{$\pm 6.2$}  & 56.2 \scriptsize{$\pm 0.5$}  \\ 
  & HEST-ccRCC & 26.6 \scriptsize{$\pm 3.8$}  & 24.0 \scriptsize{$\pm 4.0$}  & 24.3 \scriptsize{$\pm 3.3$}  & 24.9 \scriptsize{$\pm 5.4$}  & \underline{27.4} \scriptsize{$\pm 4.5$}  & 26.8 \scriptsize{$\pm 3.2$}  & \textbf{27.8} \scriptsize{$\pm 3.6$}  \\ 
  & MSI CRC (patch) & 67.5 \scriptsize{$\pm 0.0$}  & 69.1 \scriptsize{$\pm 0.0$}  & 69.0 \scriptsize{$\pm 0.1$}  & 68.0 \scriptsize{$\pm 0.0$}  & \underline{71.6} \scriptsize{$\pm 0.1$}  & 69.7 \scriptsize{$\pm 0.0$}  & \textbf{71.6} \scriptsize{$\pm 0.0$}  \\ 
  & MSI STAD (patch) & 68.6 \scriptsize{$\pm 0.0$}  & 68.6 \scriptsize{$\pm 0.0$}  & 67.4 \scriptsize{$\pm 0.1$}  & 72.4 \scriptsize{$\pm 0.0$}  & \underline{72.8} \scriptsize{$\pm 0.0$}  & 72.7 \scriptsize{$\pm 0.0$}  & \textbf{73.5} \scriptsize{$\pm 0.0$}  \\ 
 [0.05cm]\multicolumn{2}{c}{Molecular-Average} & 40.0 & 41.1 & 41.0 & 40.7 & 42.6 & \textbf{43.6} & \underline{42.8} \\ 
 \midrule 
\parbox[t]{2mm}{\multirow{9}{*}{\rotatebox[origin=c]{90}{Morphology-related}}} & Pan-cancer TIL & 92.6 \scriptsize{$\pm 0.0$}  & 92.4 \scriptsize{$\pm 0.0$}  & 91.8 \scriptsize{$\pm 0.0$}  & 91.9 \scriptsize{$\pm 0.0$}  & \underline{92.7} \scriptsize{$\pm 0.0$}  & 92.6 \scriptsize{$\pm 0.0$}  & \textbf{92.8} \scriptsize{$\pm 0.0$}  \\ 
  & TCGA Uniform (10x) & 63.9 \scriptsize{$\pm 0.0$}  & 68.3 \scriptsize{$\pm 0.0$}  & 68.7 \scriptsize{$\pm 0.0$}  & 70.2 \scriptsize{$\pm 0.0$}  & \textbf{72.9} \scriptsize{$\pm 0.0$}  & 69.9 \scriptsize{$\pm 0.0$}  & \underline{71.8} \scriptsize{$\pm 0.0$}  \\ 
  & TCGA Uniform (20x) & 69.8 \scriptsize{$\pm 0.0$}  & 67.4 \scriptsize{$\pm 0.0$}  & 67.4 \scriptsize{$\pm 0.0$}  & \textbf{77.7} \scriptsize{$\pm 0.0$}  & 71.5 \scriptsize{$\pm 0.0$}  & \underline{72.1} \scriptsize{$\pm 0.0$}  & 67.2 \scriptsize{$\pm 0.0$}  \\ 
  & BACH & 72.7 \scriptsize{$\pm 0.3$}  & 79.7 \scriptsize{$\pm 0.4$}  & 76.1 \scriptsize{$\pm 0.4$}  & 74.9 \scriptsize{$\pm 0.5$}  & \underline{88.0} \scriptsize{$\pm 0.4$}  & 75.8 \scriptsize{$\pm 1.1$}  & \textbf{93.1} \scriptsize{$\pm 0.2$}  \\ 
  & CRC-100k & 93.9 \scriptsize{$\pm 0.0$}  & 94.8 \scriptsize{$\pm 0.1$}  & 95.2 \scriptsize{$\pm 0.1$}  & 94.8 \scriptsize{$\pm 0.1$}  & \underline{96.6} \scriptsize{$\pm 0.1$}  & 95.8 \scriptsize{$\pm 0.1$}  & \textbf{97.0} \scriptsize{$\pm 0.0$}  \\ 
  & MHIST & 77.5 \scriptsize{$\pm 0.1$}  & 84.4 \scriptsize{$\pm 0.1$}  & 82.9 \scriptsize{$\pm 0.1$}  & 79.8 \scriptsize{$\pm 0.1$}  & \textbf{85.8} \scriptsize{$\pm 0.2$}  & 83.9 \scriptsize{$\pm 0.1$}  & \underline{85.2} \scriptsize{$\pm 0.1$}  \\ 
  & PCAM & 89.3 \scriptsize{$\pm 0.0$}  & 93.6 \scriptsize{$\pm 0.1$}  & \underline{94.5} \scriptsize{$\pm 0.0$}  & 94.2 \scriptsize{$\pm 0.1$}  & 93.6 \scriptsize{$\pm 0.1$}  & 94.2 \scriptsize{$\pm 0.1$}  & \textbf{94.9} \scriptsize{$\pm 0.0$}  \\ 
  & CAMELYON16 & 79.8 \scriptsize{$\pm 1.4$}  & 84.9 \scriptsize{$\pm 0.8$}  & 82.1 \scriptsize{$\pm 0.7$}  & 77.1 \scriptsize{$\pm 1.7$}  & \underline{86.5} \scriptsize{$\pm 0.2$}  & 83.2 \scriptsize{$\pm 1.4$}  & \textbf{86.8} \scriptsize{$\pm 0.4$}  \\ 
  & PANDA & 64.3 \scriptsize{$\pm 0.4$}  & \underline{69.6} \scriptsize{$\pm 0.6$}  & 66.1 \scriptsize{$\pm 0.3$}  & 68.2 \scriptsize{$\pm 0.7$}  & 65.1 \scriptsize{$\pm 0.9$}  & 67.2 \scriptsize{$\pm 0.4$}  & \textbf{70.0} \scriptsize{$\pm 0.5$}  \\ 
 [0.05cm]\multicolumn{2}{c}{Morphology-Average} & 78.2 & 81.7 & 80.5 & 81.0 & \underline{83.6} & 81.6 & \textbf{84.3} \\ 
 \midrule 
\multicolumn{2}{c}{Overall Average} & 56.4 & 58.5 & 57.9 & 57.9 & \underline{60.2} & 59.9 & \textbf{60.6} \\ 
 \bottomrule 
\end{tabular} 

\end{adjustbox}
\end{table}

\begin{table}[ht]
\caption{Model results per task, CLS+MEAN token only.
\vspace{3mm}
} 
\label{tab:cls_mean}
\centering
\begin{adjustbox}{width=1\textwidth}
\small

\begin{tabular}{p{1cm}@{}wl{2cm}wr{1.9cm}wr{1.9cm}wr{1.9cm}wr{1.9cm}wr{1.9cm}wr{1.9cm}wr{1.9cm}} 
Group & Benchmark & Phikon v2~\cite{filiot_phikon-v2_2024} & UNI~\cite{chen_towards_2024} & Gigapath~\cite{xu_whole-slide_2024} & RudolfV~\cite{dippel_rudolfv_2024}  & Virchow2~\cite{zimmermann_virchow_2024} & H-optimus-0~\cite{hoptimus0} & Atlas \\ 
\midrule 
\parbox[t]{2mm}{\multirow{12}{*}{\rotatebox[origin=c]{90}{Molecular-related}}} & HEST-COAD & 25.6 \scriptsize{$\pm 2.3$}  & 25.6 \scriptsize{$\pm 4.7$}  & \underline{30.7} \scriptsize{$\pm 0.0$}  & \textbf{31.0} \scriptsize{$\pm 2.0$}  & 25.6 \scriptsize{$\pm 3.2$}  & 30.6 \scriptsize{$\pm 0.3$}  & 29.4 \scriptsize{$\pm 1.5$}  \\ 
  & HEST-HCC & 7.8 \scriptsize{$\pm 1.2$}  & 8.3 \scriptsize{$\pm 0.5$}  & 6.8 \scriptsize{$\pm 0.4$}  & 9.4 \scriptsize{$\pm 1.7$}  & \underline{9.6} \scriptsize{$\pm 1.0$}  & 8.4 \scriptsize{$\pm 1.2$}  & \textbf{10.7} \scriptsize{$\pm 1.9$}  \\ 
  & HEST-IDC & 56.6 \scriptsize{$\pm 7.8$}  & 58.5 \scriptsize{$\pm 7.7$}  & 56.8 \scriptsize{$\pm 7.6$}  & 57.4 \scriptsize{$\pm 8.5$}  & 59.3 \scriptsize{$\pm 8.5$}  & \textbf{61.0} \scriptsize{$\pm 8.1$}  & \underline{60.4} \scriptsize{$\pm 8.3$}  \\ 
  & HEST-LUAD & 54.8 \scriptsize{$\pm 2.3$}  & 55.2 \scriptsize{$\pm 2.2$}  & 55.8 \scriptsize{$\pm 2.9$}  & \underline{57.7} \scriptsize{$\pm 1.8$}  & 56.9 \scriptsize{$\pm 1.7$}  & 57.3 \scriptsize{$\pm 2.7$}  & \textbf{58.0} \scriptsize{$\pm 1.5$}  \\ 
  & HEST-LYMPH\_IDC & 24.8 \scriptsize{$\pm 4.9$}  & 25.8 \scriptsize{$\pm 4.1$}  & 25.1 \scriptsize{$\pm 4.2$}  & 25.6 \scriptsize{$\pm 3.3$}  & 25.9 \scriptsize{$\pm 3.3$}  & \textbf{26.8} \scriptsize{$\pm 4.0$}  & \underline{26.4} \scriptsize{$\pm 4.4$}  \\ 
  & HEST-PAAD & 47.9 \scriptsize{$\pm 7.0$}  & 48.8 \scriptsize{$\pm 5.8$}  & 49.5 \scriptsize{$\pm 5.8$}  & \underline{51.1} \scriptsize{$\pm 7.8$}  & 47.3 \scriptsize{$\pm 6.9$}  & 50.9 \scriptsize{$\pm 4.3$}  & \textbf{51.8} \scriptsize{$\pm 7.4$}  \\ 
  & HEST-PRAD & 37.7 \scriptsize{$\pm 0.1$}  & 32.2 \scriptsize{$\pm 8.1$}  & \textbf{38.4} \scriptsize{$\pm 3.1$}  & 37.7 \scriptsize{$\pm 2.6$}  & 35.1 \scriptsize{$\pm 3.4$}  & 36.1 \scriptsize{$\pm 2.0$}  & \underline{38.4} \scriptsize{$\pm 0.7$}  \\ 
  & HEST-READ & 18.5 \scriptsize{$\pm 5.9$}  & 17.4 \scriptsize{$\pm 6.3$}  & 18.8 \scriptsize{$\pm 6.3$}  & 19.9 \scriptsize{$\pm 6.8$}  & 21.1 \scriptsize{$\pm 5.0$}  & \textbf{24.1} \scriptsize{$\pm 2.7$}  & \underline{22.8} \scriptsize{$\pm 3.1$}  \\ 
  & HEST-SKCM & 58.4 \scriptsize{$\pm 6.2$}  & 63.0 \scriptsize{$\pm 2.4$}  & 58.8 \scriptsize{$\pm 5.8$}  & 61.8 \scriptsize{$\pm 4.6$}  & \underline{63.7} \scriptsize{$\pm 3.1$}  & \textbf{66.1} \scriptsize{$\pm 5.8$}  & 62.5 \scriptsize{$\pm 2.4$}  \\ 
  & HEST-ccRCC & 27.3 \scriptsize{$\pm 4.0$}  & 25.3 \scriptsize{$\pm 3.8$}  & 24.9 \scriptsize{$\pm 4.1$}  & 25.3 \scriptsize{$\pm 5.2$}  & 27.4 \scriptsize{$\pm 5.4$}  & \underline{29.0} \scriptsize{$\pm 3.8$}  & \textbf{29.4} \scriptsize{$\pm 5.5$}  \\ 
  & MSI CRC (patch) & 68.8 \scriptsize{$\pm 0.1$}  & 69.5 \scriptsize{$\pm 0.0$}  & 70.4 \scriptsize{$\pm 0.1$}  & 69.9 \scriptsize{$\pm 0.1$}  & \textbf{74.0} \scriptsize{$\pm 0.0$}  & 71.2 \scriptsize{$\pm 0.1$}  & \underline{73.6} \scriptsize{$\pm 0.0$}  \\ 
  & MSI STAD (patch) & 71.2 \scriptsize{$\pm 0.1$}  & 70.5 \scriptsize{$\pm 0.0$}  & 71.0 \scriptsize{$\pm 0.1$}  & 74.1 \scriptsize{$\pm 0.1$}  & \underline{74.8} \scriptsize{$\pm 0.2$}  & 73.6 \scriptsize{$\pm 0.0$}  & \textbf{76.0} \scriptsize{$\pm 0.1$}  \\ 
 [0.05cm]\multicolumn{2}{c}{Molecular-Average} & 41.6 & 41.7 & 42.3 & 43.4 & 43.4 & \underline{44.6} & \textbf{44.9} \\ 
 \midrule 
\parbox[t]{2mm}{\multirow{9}{*}{\rotatebox[origin=c]{90}{Morphology-related}}} & Pan-cancer TIL & 92.9 \scriptsize{$\pm 0.0$}  & 92.6 \scriptsize{$\pm 0.0$}  & 92.3 \scriptsize{$\pm 0.0$}  & 92.6 \scriptsize{$\pm 0.1$}  & \textbf{93.1} \scriptsize{$\pm 0.1$}  & \underline{93.0} \scriptsize{$\pm 0.1$}  & 93.0 \scriptsize{$\pm 0.0$}  \\ 
  & TCGA Uniform (10x) & 64.0 \scriptsize{$\pm 0.0$}  & 68.6 \scriptsize{$\pm 0.0$}  & 69.1 \scriptsize{$\pm 0.0$}  & 70.6 \scriptsize{$\pm 0.0$}  & \textbf{73.0} \scriptsize{$\pm 0.0$}  & 70.4 \scriptsize{$\pm 0.0$}  & \underline{71.7} \scriptsize{$\pm 0.0$}  \\ 
  & TCGA Uniform (20x) & 69.4 \scriptsize{$\pm 0.0$}  & 67.8 \scriptsize{$\pm 0.0$}  & 68.0 \scriptsize{$\pm 0.0$}  & \textbf{78.1} \scriptsize{$\pm 0.0$}  & 71.5 \scriptsize{$\pm 0.0$}  & \underline{72.4} \scriptsize{$\pm 0.0$}  & 67.8 \scriptsize{$\pm 0.0$}  \\ 
  & BACH & 73.8 \scriptsize{$\pm 0.3$}  & 80.1 \scriptsize{$\pm 1.0$}  & 80.2 \scriptsize{$\pm 0.3$}  & 76.9 \scriptsize{$\pm 0.3$}  & \underline{88.7} \scriptsize{$\pm 0.3$}  & 74.2 \scriptsize{$\pm 0.5$}  & \textbf{92.5} \scriptsize{$\pm 0.5$}  \\ 
  & CRC-100k & 95.5 \scriptsize{$\pm 0.0$}  & 95.4 \scriptsize{$\pm 0.2$}  & 95.9 \scriptsize{$\pm 0.0$}  & 96.0 \scriptsize{$\pm 0.1$}  & \underline{96.7} \scriptsize{$\pm 0.1$}  & 96.2 \scriptsize{$\pm 0.1$}  & \textbf{97.1} \scriptsize{$\pm 0.1$}  \\ 
  & MHIST & 78.4 \scriptsize{$\pm 0.4$}  & 84.0 \scriptsize{$\pm 0.1$}  & 83.1 \scriptsize{$\pm 0.1$}  & 80.5 \scriptsize{$\pm 0.0$}  & \underline{85.9} \scriptsize{$\pm 0.1$}  & 85.0 \scriptsize{$\pm 0.2$}  & \textbf{86.4} \scriptsize{$\pm 0.1$}  \\ 
  & PCAM & 90.0 \scriptsize{$\pm 0.1$}  & 93.6 \scriptsize{$\pm 0.1$}  & 94.3 \scriptsize{$\pm 0.1$}  & \underline{94.6} \scriptsize{$\pm 0.1$}  & 93.9 \scriptsize{$\pm 0.1$}  & 94.3 \scriptsize{$\pm 0.1$}  & \textbf{94.8} \scriptsize{$\pm 0.0$}  \\ 
  & CAMELYON16 & 79.1 \scriptsize{$\pm 1.2$}  & 85.0 \scriptsize{$\pm 0.5$}  & 80.6 \scriptsize{$\pm 0.7$}  & 76.3 \scriptsize{$\pm 0.5$}  & \underline{86.0} \scriptsize{$\pm 0.3$}  & 84.0 \scriptsize{$\pm 0.8$}  & \textbf{86.7} \scriptsize{$\pm 0.6$}  \\ 
  & PANDA & 65.3 \scriptsize{$\pm 0.4$}  & 69.0 \scriptsize{$\pm 0.5$}  & 69.6 \scriptsize{$\pm 0.7$}  & \underline{69.6} \scriptsize{$\pm 0.4$}  & 66.4 \scriptsize{$\pm 1.1$}  & 68.0 \scriptsize{$\pm 0.6$}  & \textbf{70.5} \scriptsize{$\pm 0.5$}  \\ 
 [0.05cm]\multicolumn{2}{c}{Morphology-Average} & 78.7 & 81.8 & 81.5 & 81.7 & \underline{83.9} & 81.9 & \textbf{84.5} \\ 
 \midrule 
\multicolumn{2}{c}{Overall Average} & 57.5 & 58.9 & 59.1 & 59.8 & \underline{60.8} & 60.6 & \textbf{61.9} \\ 
 \bottomrule 
\end{tabular} 

\end{adjustbox}
\end{table}


\end{document}